\documentclass[letterpaper]{article}
\usepackage{iccc}

\usepackage{graphicx}
\usepackage{times}
\usepackage{helvet}
\usepackage{courier}
\usepackage{hyperref}
\title{Learning to Blend Computer Game Levels \\
}
\author{Matthew Guzdial, Mark Riedl\\
Entertainment Intelligence Lab, School of Interactive Computing\\
Georgia Institute of Technology\\
Atlanta, GA 303  USA\\
mguzdial3@gatech.edu, riedl@cc.gatech.edu\
}
\begin{document} 
\maketitle
\begin{abstract}
\begin{quote}
We present an approach to generate novel computer game levels that blend different game concepts in an unsupervised fashion. Our primary contribution is an analogical reasoning process to construct blends between level design models learned from gameplay videos. The models represent probabilistic relationships between elements in the game. An analogical reasoning process maps features between two models to produce blended models that can then generate new level chunks. As a proof-of-concept we train our system on the classic platformer game Super Mario Bros. due to its highly-regarded and well understood level design. We evaluate the extent to which the models represent stylistic level design knowledge and demonstrate the ability of our system to explain levels that were blended by human expert designers.
\end{quote}
\end{abstract}

\section{Introduction}

Concept blending is a powerful tool for problem solving in which two independent solutions combine into a novel solution referred to as a \textit{blend}. It has been presented as a fundamental cognitive process and linked to the creation of creative artifacts (e.g. a griffin can be described as a blend between a lion and a bird)\cite{fauconnier2008way}. Concept blending has traditionally appeared in expert systems applications, where a human expert encodes concepts from a particular field such as architecture, engineering, or mathematics \cite{goel1997design,bou2015role}. Despite the process' creative potential, it has not appeared in the domain of video games to any large extent, even though games are well-suited to explorations of computational creativity \cite{liapis2014computational}. This is likely due to concept blending ---and many other computational creativity techniques--- relying on high quality knowledge bases. The quality of this ``knowledge base" determines the quality of the blends a system is capable of constructing, meaning that a human expert often has to iterate upon the concepts in a knowledge base multiple times. In addition to the knowledge base, many concept blending systems require a means of evaluating blends, requiring human-authored heuristics. 

Concept blending systems take a significant amount of human effort to construct. Machine learning could in theory derive a knowledge base from a corpus of examples, thus reducing the requirement of human input. However, knowledge learned from machine learning techniques tends to be noisy, full of inconsistencies and mistakes that could thwart typical approaches to concept blending.

We present an unsupervised approach to concept blending video game levels, informed by a knowledge base learned from gameplay videos. The use of gameplay video is key to the unsupervised nature of our system as the system can infer human knowledge for an exemplar game without requiring explicit human authoring. The learned knowledge base takes the form of probabilistic graphical models that are robust to the noisiness of machine learning with sufficient data. The models learn the likelihood of relationships between level elements, and can therefore evaluate the relative likelihood of a level, meaning that the blended models can evaluate blends without a human authored heuristic. We make use of Super Mario Bros. as a proof-of-concept game for our system, due to its popularity and highly regarded level design.

Our contributions are as follows: (1) a novel concept blending approach to blend models capable of generation and evaluation, (2) a human evaluation of our system's ability to evaluate how stylistically similar an input level is to exemplar gameplay levels, and (3) a case study of our blended models' evaluation of human expert blended levels.

\section{Background}
Fauconnier and Turner \shortcite{fauconnier1998conceptual} formalized the ``four space" theory of concept blending. In this theory they described four spaces that make up a blend: the unblended solutions are the two \textit{input spaces}, points from both input spaces are projected into a common \textit{generic space} to identify equivalent points, and these equivalent points are projected into a \textit{blend space}. In the blend space, novel structure and patterns arise from the projection of equivalent points, leading to discovering creative, novel solutions. Fauconnier and Turner \shortcite{fauconnier1998conceptual,fauconnier2008way} argued this was a ubiquitous process, occurring in discourse, problem solving, and general meaning making. 

Concept blending systems tend to follow some variation of the four spaces theory, but there exists a great variety of techniques to map between the concepts present in the various spaces \cite{falkenhainer1989structure}. Analogical reasoning has traditionally been one of the leading conceptual mapping approaches, as it maps concepts based on relative structure instead of surface features. This type of structural mapping has proven popular as it tends to better match human problem solving  \cite{goel2015biologically,bou2015role}. However, such analogical reasoning systems require a non-trivial amount of human input, as a human author most encode concepts in terms of their structure and how to compare structural information within a domain. 

Due to the large amount of authorship required, its unclear if the creative output of such a system arises from concept blending algorithms or human's creativity when encoding structures. Recently O'Donoghue et al. \shortcite{o2015stimulating} have looked into deriving this knowledge automatically from text corpora, producing graphical representations of nodes and their verb connections. Our own work runs parallel to O'Donoghue et al., but in the domain of two dimensional video games levels and without the dependency rules that exist in the english language. 

Concept blending, based on analogy or any other mapping technique, is not commonly used in video games. Prior work has looked into knowledge intensive concept blending systems to create new elements of video games such as sound effects and 3D models \cite{ribeiro2003model,martins2004enhancing}. The Game-O-Matic system made use of concept mapping to match verbs onto game mechanics to create arcade-style games based on human-authored mapping knowledge \cite{treanor2012game}. Gow and Corneli \shortcite{gow2015towards} proposed a system to generate small games via amalgamation \cite{ontanon2010amalgams}. Permar and Magerko \shortcite{permar2013conceptual} presented a system to produce novel interactive narrative scripts via concept blending, using analogical processing. Permar and Magerko's work is similar to our own, in that these scripts can be understood as the equivalent of ``levels" for interactive narrative, but differ in their use of human-authored scripts rather than learning from a corpus of exemplars. In addition the work presented in this paper focuses on a two-dimensional platformer game, a significantly more complex domain to model than interactive narrative.

Our work is inspired by probabilistic graphical models from the computer graphics field that encode style from scene and object exemplars \cite{kalogerakis:2012:SIG,guerrero:2015:SIG,emelien:2015:SIG}. In these approaches, 3D scenes are broken into individual objects and parts, with each part and important relationships tagged by a human expert. Categories of these tagged exemplars are then used to train a probabilistic graphical model, representing style as the probability of seeing certain object pairs and their relative relationships. Our approach thus avoids much of the human effort of these systems: categorizing the exemplar input via a clustering technique, tagging individual elements via machine vision, and probabilistically determining important relationships rather than explicitly encoding them. In terms of concept blending, there has been work in blending individual tagged exemplars together based on surface level features of components \cite{alhashim2014topology}. Our work focuses on blending the models learned from exemplars rather than individual exemplars, and makes use of structural information for concept mapping.
\begin{figure*}[t]
  \centering
  \includegraphics[height=2.1in]{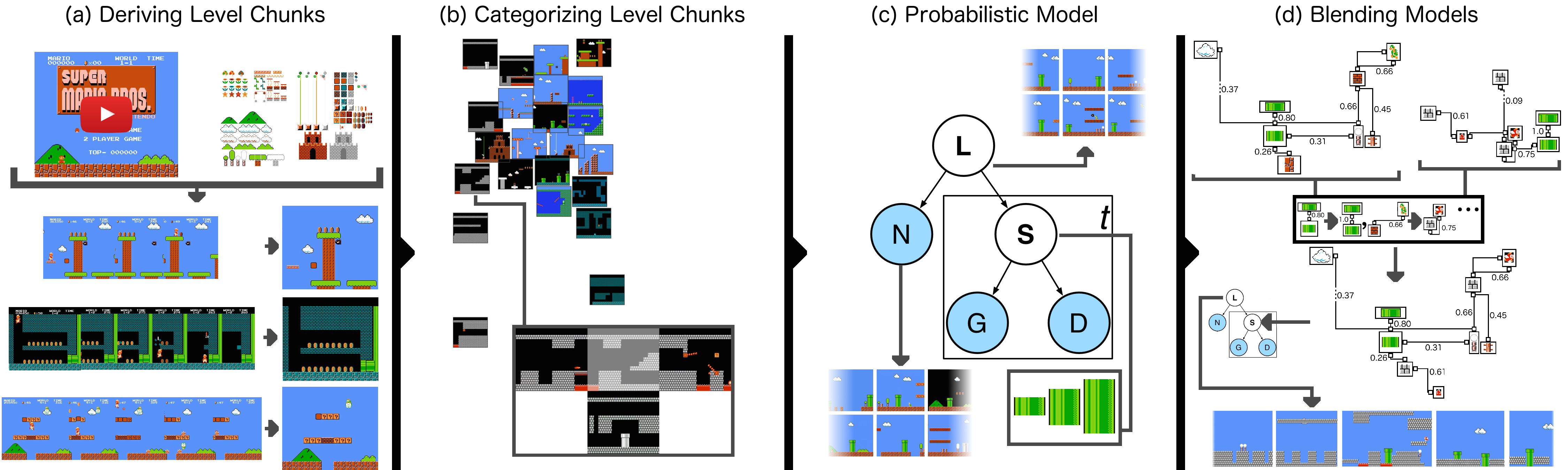}
  \caption{Visualization of the entire process of model building.}
  \label{fig:SystemOverview}
\end{figure*}
\section{System Overview}
The goal of our work is to develop a computational system capable of generating novel game levels by blending different concepts from the game together. For example, we may wish to generate a level of Super Mario Bros. in which Mario swims through an underwater castle. Our system as a whole can be understood as containing three parts, operating sequentially. First our system automatically derives sections of levels from gameplay video and categorizes these sections according to their features. Second, the system derives probabilistic graphical models from each category. At this point in the process, our system can be used to generate game levels that closely resemble, but are different, from existing game levels \cite{2016arXiv160207721G}. Lastly, our system can blend these learned models together using structural information to produce a final model that can produce creative, novel game levels. We chose the highly regarded, classic platformer Super Mario Bros. to test our approach.

We begin by supplying our system with two things: a set of videos to learn from and a sprite palette as seen in the top left corner of Figure \ref{fig:SystemOverview}. This input is simple to produce with the advent of ``Let's Plays" and ``Long Plays". By sprite palette we indicate the set of ``sprites" or individual images used to build levels of a 2D game. For this proof-of-concept we found nine videos representing entire playthroughs of Super Mario Bros. and a fan-authored spritesheet. With these elements the system makes use of OpenCV, an open-source machine vision toolkit, to determine the number and placement of sprites in each frame of the video \cite{Pulli:2012:RCV:2184319.2184337}. It then combines frames into \textit{level chunks}, the actual geometry that a frame sequence represents. Level chunks include both the sprite geometry and the length of time the player stays in that chunk. These chunks are then clustered into categories of chunk types as seen in Figure \ref{fig:SystemOverview}b.

Each learned level chunk category is used as the basis for training a probabilistic model, visualized in Figure \ref{fig:SystemOverview}c. The system learns what possible sprite shape ``styles" exist in a given category of level chunk, and the probability of relative positions between these shapes. This probabilistic approach makes up for the imperfect nature of machine vision, as mistakes disappear with sufficient data. These learned models are very large, and so the system generates an abstracted graph called an S-structure graph for blending as seen at the top of Figure \ref{fig:SystemOverview}d. The structure between sprite shape styles are then mapped from one S-structure graph to another in order to conceptually map elements from one model onto another. These mappings are then used to transform the lower-level, more detailed model into a blended model.

\section{Model Learning}
Our system learns a generative, probabilistic model of shape to shape relationships from gameplay videos. Given this paper's focus on blending we give a brief description of the model learning process here, for further detail please see \cite{2016arXiv160207721G}. These types of probabilistic graphical models, common in the object and scene modeling field, require a set of similar exemplars as input. These sets are typically categories of 3D models, decided on by a human expert. Given that the input to our system is gameplay video, we must determine (1) what input a probabilistic model should learn from and (2) how to categorize this input in an unsupervised fashion to ensure the required similarity. For the input to our system we define the level chunk, a short segment of a level. For the categorization we make use of K means clustering with K estimated with the distortion ratio \cite{Pham01012005}. Each category is then used as input to learn a generative, probabilistic model. 
\begin{figure*}[ht]
  \centering
  \includegraphics[height=1.2in]{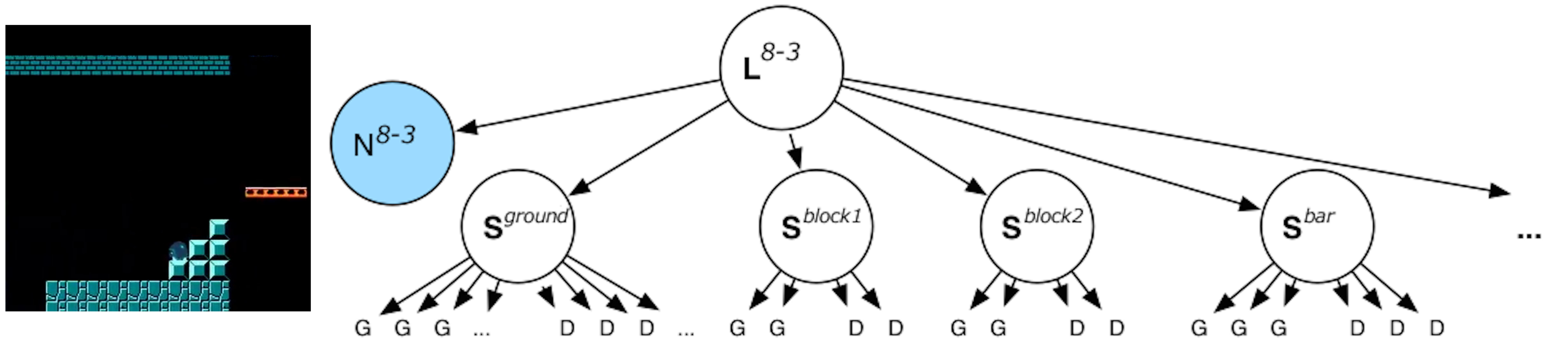}
  \caption{Visualization of a final L Node and one of the example chunks used to train it.}
  \label{fig:finalModel}
\end{figure*}
\subsection{Probabilistic Model}
The system builds a probabilistic graphical model from each of the level chunk categories. The intuition for this per-category learning is that different types of level chunks have different relationships, and therefore different models must be learned on an individual category basis. The model extracts values for latent variables to represent probabilistic design rules of a level chunk category. Figure \ref{fig:SystemOverview}c contains a visual representation of the probabilistic model, along with visualizations of three node types. White nodes represent hidden variables, with the blue node values derived directly from the level chunks in a category. Figure \ref{fig:finalModel} represents a final learned model for an individual category (category ``8-3", the third cluster found after reclustering cluster eight), along with a representative level chunk. 
\begin{figure}[tb]
  \centering
  \includegraphics[width=2.5in]{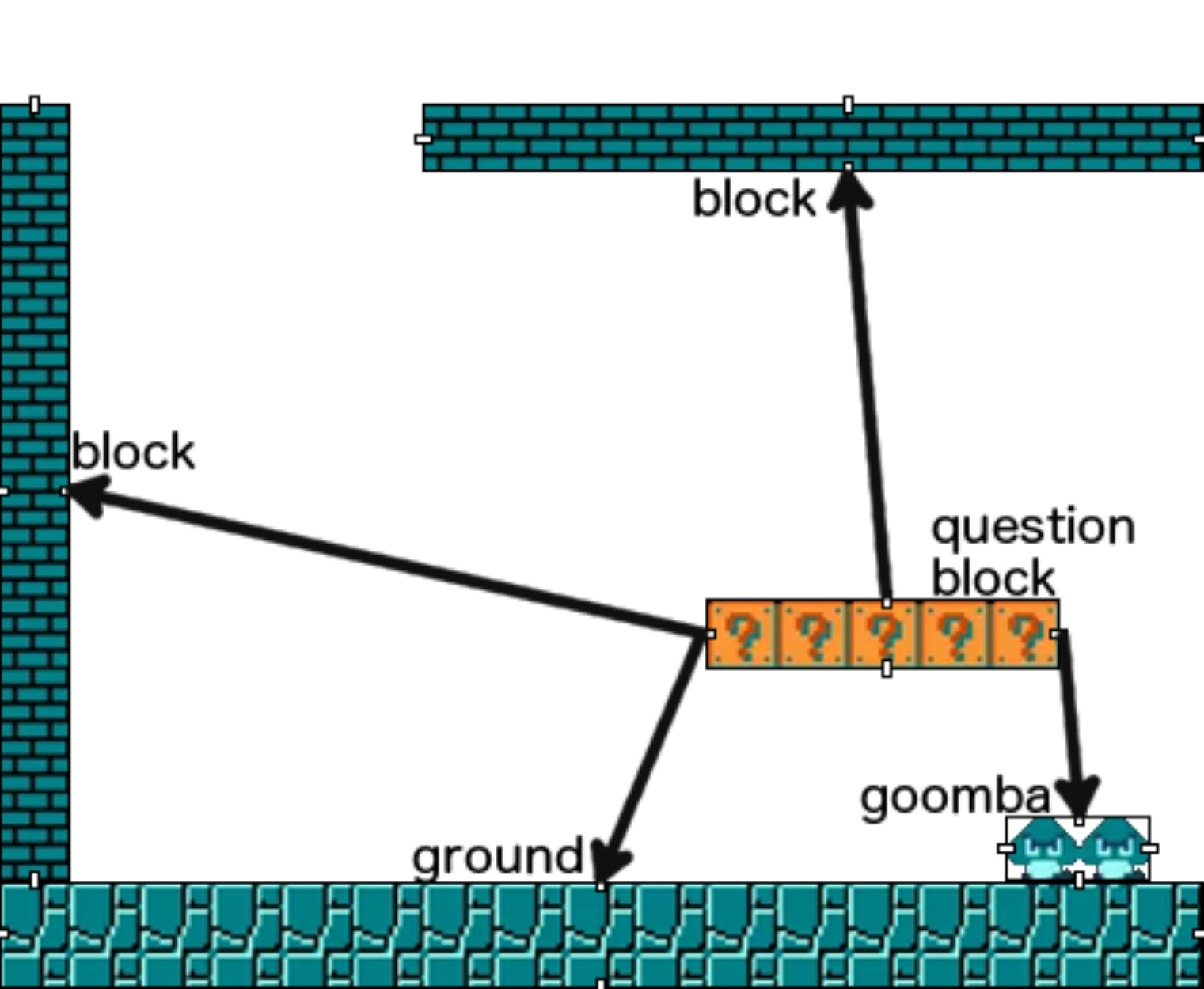}
  \caption{Example of a D Node}
  \label{fig:dNode}
\end{figure}

The three observable nodes are the G node, D node, and N node. The G node represents the sprite ``geometry", an individual sprite shape of sprite type $t$. Sprite shapes in this case are built by connecting all adjacent sprites of the same type $t$ (e.g. ground, block, coin). These shapes can differ considerably, Figure \ref{fig:dNode} contains two ``block" shapes differing in both orientation and size. The D node represents the set of all relative relationships between a given G node and all other G nodes in it's level chunk. The D node in Figure \ref{fig:dNode} is the set of vectors capturing relative orientation and direction between the question block shape and all other G nodes in the chunk (two block shapes, one goomba shape, and one ground shape). The vectors connect at the cardinal points in order to better represent symmetry in the design. Each D node is paired to a specific G node, as in Figure \ref{fig:dNode} that visualizes the question block shape's D node. The N node is the last directly observed variable. It represents the number of individual atomic sprite values in a particular level chunk. In the case of Figure \ref{fig:dNode} there are two goombas, seventeen ground sprites, etc. 
\begin{figure*}[tb]
  \centering
  \includegraphics[height=0.9in]{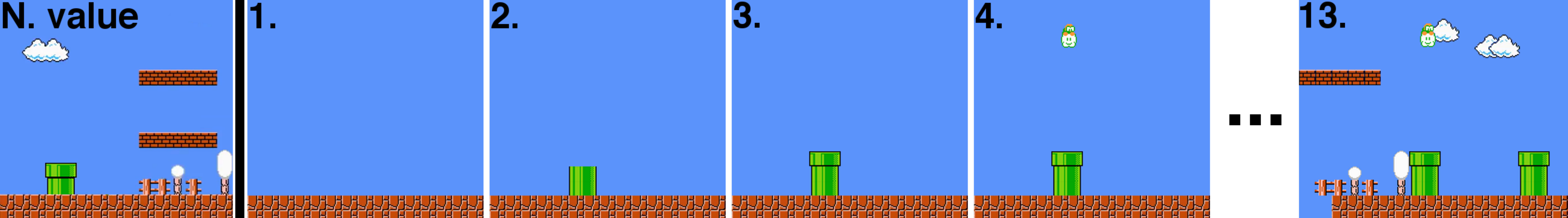}
  \caption{A visualization of the chunk generation process, beginning with an N node value and a single ``ground" shape. }
  \label{fig:generationSteps}
\end{figure*}

The first latent variable is the S node, it represents ``styles" of sprite types. These styles can vary either in geometry or relative position. For example, there are a variety of possible arrangements and positions of pipe bodies, as seen in the lower right of Figure \ref{fig:SystemOverview}c. They can come in groups ranging in \textit{size} from one to four, and can differ in \textit{position}, appearing on top of the ground, on stairs, or out of the bottom of the screen. The system learns the values and number of S nodes by clustering G and D node pairs. By pairs of G and D nodes we mean that each shape is paired with the set of connections from it to everything in it's chunk. This process is accomplished by sprite type, meaning that there is at least one S node for each type of sprite. With a fully formed S node we can now determine the probability of an S node shape of a specific type at a given relative distance, given another S node shape. More formally: $P(g_{s1},r_d | g_{s2})$ or the probability of a G node from \textit{within} a particular S node, given a relative distance to a second G node. For example in Figure \ref{fig:dNode}, goomba shapes have a high probability of co-occurring with ground shapes at those same relative positions.

The L Node represents a specific style of level chunk, the intuition behind it is that it is constituted by the different styles of sprite shapes (S) and the different kinds of chunks that can be built with those shapes (N). Once again the system represents this as a clustering problem, this time of S nodes. Each S node tracks the N node values that arose from the same chunk as it's G and D nodes. Essentially, each S node knows the level chunks from the original Mario that represented its ``style" of shape. Figure \ref{fig:finalModel} represents a final learned L Node and all of it's children. Notice the multiple S nodes of the ``block" type, with the singular ``ground" S node.

\subsection{Generation of Novel Level Chunks}
L nodes can be used to generate novel level chunks. The generation process is a simple greedy search algorithm, attempting to maximize the following scoring function: 
\begin{equation}
1/N * \sum_{i=1}^{N}\sum_{j=1}^{N} p(g_i | g_j, r_{i-j})
\end{equation}
Where $N$ is equal to the current number of shapes in a level chunk, $g_i$ is the shape at the $i$th index, $g_j$ is the shape at the jth index, and $r_{i-j}$ is the relative position of $g_i$ from $g_j$. This is equivalent to the average of the probabilities of each shape in terms of its relative position to every other shape. 

The generation process begins with two things: a single shape chosen randomly from the space of possible shapes in an L node, and a random N node value to serve as an end condition. The N nodes hold count data of sprites from the original level chunks in a category. For example in Figure \ref{fig:generationSteps} the top image is a level chunk that informed an N node value with: ``blocks: 10", ``pipeTop: 1" and so forth. This N node value can therefore serve as an end condition to the process as it can specify how many of each sprite type a generated chunk needs to be complete.

In every step of the generation process, the system creates a list of possible next shapes, and tests each, choosing the one that maximizes its scoring function. These possible next shapes are chosen according to two metrics: (1) shapes that are still needed to reach the N node value-defined end state and (2) shapes that are required given a shape already in the level chunk. For example in Figure \ref{fig:generationSteps} from step 1 to step 2 the ``pipeBody" shape is added in order to get closer to the end state, while from step 3 to step 4 the ``lakitu" enemy is added as the system deems it to be required with the style of pipeTop shape added in step 3. The system defines a shape to require another shape if $p(s_1 | s_2)>0.95$, or if the two shapes co-occur more than 95\% of the time. The process ends either because the chunk reaches a sufficient number of sprites as determined by the N node, or the probability of adding any further shapes is too low ($p<0.05$).
\begin{figure}[tb]
  \centering
  \includegraphics[width=3.2in]{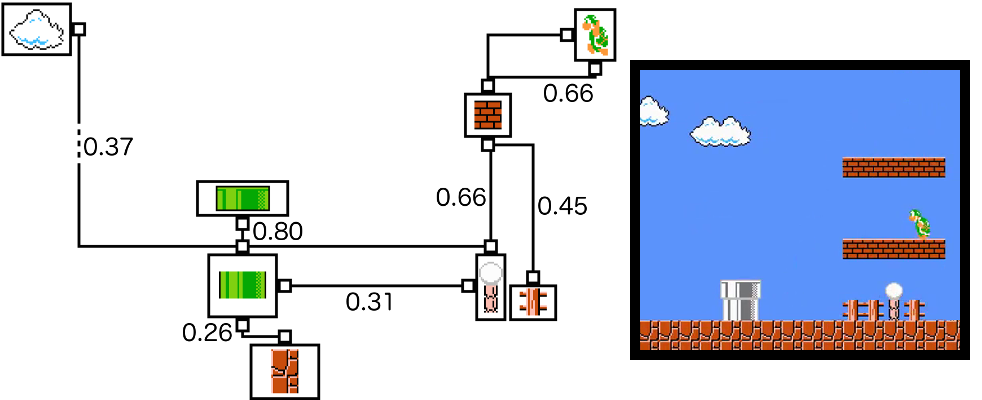}
  \caption{S-structure graph and an example of its instantiation.}
  \label{fig:abstractGraph}
\end{figure}
\section{Blending}
The levels generated from a learned probabilistic model tend to resemble the original Super Mario Bros. levels, and while novel, may not be considered creative or surprising. Concept blending serves as a well-regarded approach to produce creative artifacts, but the learned models extracted from gameplay videos are not suited to concept mapping. Instead of using these models directly, our system takes the common concept blending approach and transforms our detailed model into a more abstract model in order to find mappings \cite{goel1997design}. We define this \textit{S-structure graph} as the set of S nodes, styles of sprite shapes in a model, and a set of edges representing probabilistic relative positions between them as seen in Figure \ref{fig:abstractGraph}. In most concept blending systems the abstraction knowledge (e.g. a door and a cabinet are both ``openable furniture"") is encoded by a human expert. Instead we can make use of the learned probabilistic relationships.

Figure \ref{fig:abstractGraph} gives an example of a final S-structure graph on the left derived from an L Node trained on level chunks like that on the right. Each box and image represents an S node, the lines between them are D node connections, vectors connecting the cardinal points of the shape styles. The D node connections also have a probability [0...1] corresponding to how likely they are to appear. The S-structure graphs form the basis of structural comparisons between different types of level chunks.

Each S Node has many more D node connections than are used in the S-structure graph. The system uses a subsection of connections equal to the minimum number of connections with the maximum probability to creates a fully-connected graph. The system defines a threshold $\Theta_s$ for each S node, with a starting value of 1.0. The system decreases this value iteratively for the current most unconnected node, then adding all the connections of equal or greater probability than $\Theta_s$ for each S node to a potential graph. When the graph is fully connected, the process stops. 

Concept blending systems typically have a concept of a \textit{source} space and \textit{target} space. Our approach is the same, in that an L node to blend from (source) and an L node to blend to (target) must be selected. Each relationship ---D node connection--- from the source graph is mapped to the closest relationship on the target graph. This mapping is a simple closest match, based on a function that equally weights differences in probability with the cosine distance. This list of D node connection mappings can be transformed into a list of S node mappings via referencing the S nodes the relationship exists between. The structural mapping between these relationships therefore serves as a basis for potential S node mappings, with the final S node mappings determined according to the greatest evidence and the target of the blend.

Consider two mapped D node connections from two different S-structure graphs, one representing the relationship between ``ground" and ``goombas" and the other the relationship between ``sea blocks" and ``squids". From these the system can derive the mappings ``ground to sea blocks" and ``goombas to squids". The system then takes the final mappings with the greatest evidence. Rather than map \textit{all} of the S nodes from one probabilistic model to another, the system can specify a \textit{target} for the blend, a desired final set of S nodes, and the system can choose only the mappings that fit this final set. For example, if our desired final set was ``sea blocks, squids, and goombas" then the system could accept the mapping ground to sea blocks, but not goombas to squids. This final mapping is then used to transform the \textit{source} L node, which means changing N, and S node values within the L node. For example, if previously there existed a relationship between goomba and ground, there would now exist a relationship between goomba and seablock. 

\section{Evaluation}
In this section we present results from two distinct evaluations meant to demonstrate the utility of our system. The first evaluation is a human study that demonstrates that our probabilistic graphical model captures humans' intuitions of level design style. The second evaluation is a case study, demonstrating that our system's blended models can explain human-created expert blends significantly better than when the system is not allowed to blend models.
\begin{table}[tb]
  \centering
  \caption{The results of comparing our system's rankings and participant rankings per question.}
  \begin{tabular}{|c|c|c|}
    \hline
    Category & $r_s$ & $p$\\
     \hline
    Style & \textbf{0.6115095} & \textbf{2.2e-16}\\ 
    \hline
     Design & \textbf{0.51948} & \textbf{2.2e-16}\\
    \hline
    Fun &  \textbf{0.2729658} & \textbf{3.745e-5}\\
    \hline
    Frustration &  \textbf{-0.4393904} & \textbf{6.79e-12}\\
    \hline
    Challenge & \textbf{-0.387222} & \textbf{2.351e-09}\\
    \hline
    Creativity & -0.1559725 & \textbf{0.02007}\\
    \hline
  \end{tabular}
\end{table}
\subsection{Model Evaluation}
The first evaluation shows that the models learned by our system capture human design intuition. We do this by showing that Equation 1 scores Super Mario Bros. levels similarly to humans.

We ran a human subjects study in order to obtain human level rankings to compare to our system rankings. In the study, individuals played through a series of three levels in the vein of Super Mario Bros., the first of which was always a level from the original Super Mario Bros., while the other two levels were chosen randomly from a set of fifteen novel levels. The fifteen novel levels were generated from three generators: the Snodgrass and Onta{\~n}{\'o}n \shortcite{snodgrass2014experiments} generator, the Dahlskog and Togelius \shortcite{dahlskog} generator, and our own generative system. After playing all three levels subjects were asked to rank the three levels they played on measures of style (defined as more ``mario-like"), design, fun, frustration, challenge, and creativity. If our hypothesis is correct, we'd expect to see the human ranking of levels correlate strongly with our system's predicted rankings of these levels based on our system's level chunk scoring function.

In order to use the scoring function in equation 1 for entire levels we broke each into chunks of uniform length, randomly selected from these chunks to ensure equally sized distributions, and then used the maximum scoring L node to score each chunk. This gave a distribution of scores over an entire level, and we then determined an absolute ranking of levels according to the median values of these distributions. For any trio of levels a human participant ranked our system could then determine its own predicted rankings.

We ran this study for two months and collected seventy-five respondents. We compared the participant rankings and our system's predicted rankings with Spearman's Rank-Order Correlation. Table 1 summarizes the results with significant p-values and correlations in bold.
\begin{figure}[tb]
  \centering
  \includegraphics[width=3.3in]{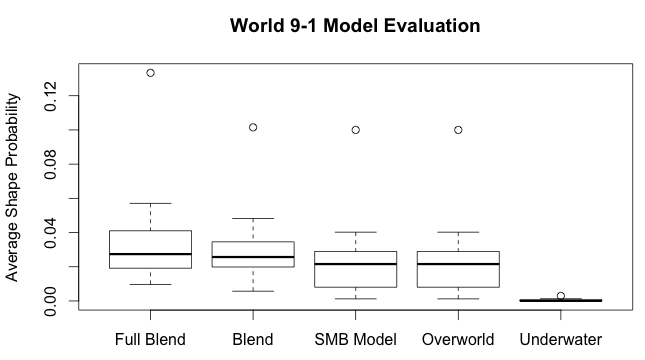}
  \caption{Distributions of scores based on evaluating World 9-1 with various models.}
  \label{fig:world91Box}
\end{figure}
\begin{figure}[tb]
  \centering
  \includegraphics[width=3.3in]{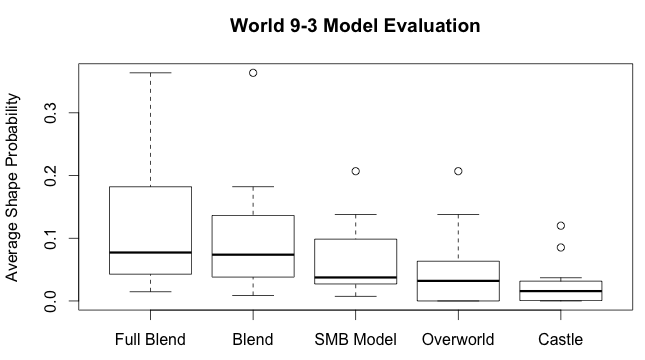}
  \caption{Distributions of scores based on evaluating World 9-3 with various models.}
  \label{fig:world93Box}
\end{figure}

The strongest correlation present is for the style rankings, which provides strong evidence that our model captures stylistic information. The other correlations can be explained as a side-effect of our model training on the well-designed Super Mario Bros. levels. The very weak correlation between the creativity rankings and our system's rankings is likely due to the lack of a strong cultural definition of creativity in video game levels. The respondent ranking distributions on a per-generator basis did not differ significantly, further suggesting that this interpretation is accurate, as otherwise we'd expect to see some generators creating more ``creative" levels than others.
\begin{figure*}[tb]
  \centering
  \includegraphics[height=0.42in]{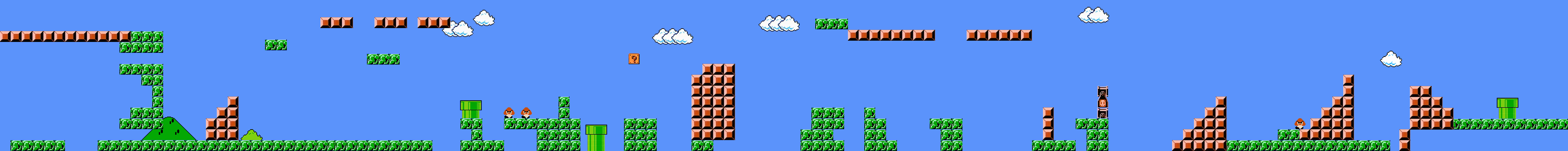}
  \caption{A high-quality blended level according to the model trained off of World 9-1.}
  \label{fig:uwLvl}
\end{figure*}
\begin{figure*}[tb]
  \centering
  \includegraphics[height=0.42in]{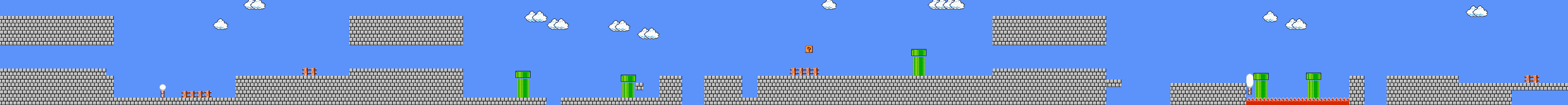}
  \caption{A high-quality blended level according to the model trained off of World 9-3.}
  \label{fig:high}
\end{figure*}
\subsection{Blending Evaluation: Lost Levels}
The evaluation of blending techniques is a traditionally difficult problem due to the subjective nature of blend quality. Given that our blended models are generative, we could run a human study on levels generated from these models. However, our initial human study demonstrated that human subjects do not tend to agree on the creativity of a level, indicating that this type of study would be inconclusive. Learned models can also be used to evaluate. An alternative way to determine the quality of our blended models is to ascertain how well they account for human-expert blends.

In the case of Super Mario Bros., the designer Shigeru Miyamoto designed a second game known as Super Mario Bros.: Lost Worlds based on the original game. This game included a ``fantasy world" in which Miyamoto added a series of more whimsical levels. These include two levels that can be understood as blends of Super Mario Bros. level types.\footnote{\url{http://www.mariowiki.com/World_9_(Super_Mario_Bros.:_The_Lost_Levels)}}  Level 9-1 uses a combination of sprites found otherwise only separately in ``underwater" and ``overworld" levels. The level includes castles, clouds, and bushes that only appear in overworld levels appearing with coral and squids. Level 9-3 on the other hand uses a combination of sprites otherwise found only separately in ``castle" and ``overworld" levels. The level includes elements from overworld levels alongside lava and castle walls. Due to their ``blended" nature, we hypothesize that our blending technique can create models that explain these human blends significantly better than our original, unblended models trained on the Super Mario Bros. levels. That is, how well do the actual relationships between sprites in Lost Levels match the predicted relationships in our models. To rank these levels with our system we used the same strategy as our earlier model evaluation, sectioning off each level into uniform chunks and evaluated each chunk with a set of learned models.

We created four different versions of our system to create four distinct types of learned model:  
\begin{itemize}  
\item \textbf{SMB Model}: The Super Mario Bros. (SMB) model represents the set of L nodes learned from gameplay video of the original game.
\item \textbf{Blended Model}: To construct a blended model the system first chooses what of the original L nodes to blend. The system constructs this initial set by choosing the L node that maximally explains each uniform chunk of the blended level. The system then blends each each pair of L nodes in the set as both the source and target L node using the blended level as the target for the blend. This model can be thought of as an unsupervised model, and represents the ideal interpretation of our approach.
\item \textbf{Level Type Model}: We constructed additional models via hand-tagging each of L node with it's level type. For example, ``Overworld" to represent the above ground levels, ``Underwater" and ``Castle". These models represent subsections of the larger SMB model. We parsed each blended level with the level type models that made up its blend. World 9-3 (Figure \ref{fig:world93}) was therefore parsed with the ``Overworld" and ``Castle" models. 
\item \textbf{Full Blended Model}: We constructed the largest possible blended model for each level as a ``full" blended model. We constructed this model by taking all of the L nodes tagged with the two level types for each blended level, and blending all of the L nodes together for all possible pairs, leading to a massive final blended model. This model served as an upper-bound of performance for our blending technique given human knowledge of level types, and can therefore be considered a supervised model.
\end{itemize}
\begin{figure*}[tb]
  \centering
  \includegraphics[height=0.42in]{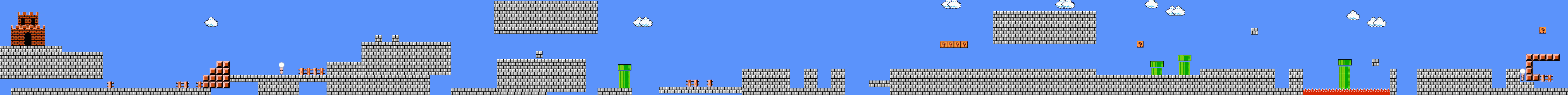}
  \caption{A lower quality blended level according to the model trained off of World 9-3.}
  \label{fig:mid}
\end{figure*}
\begin{figure*}[tb]
  \centering
  \includegraphics[height=0.42in]{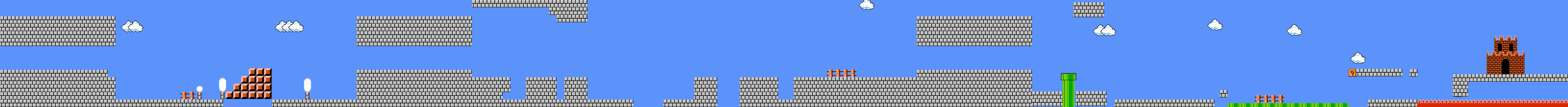}
  \caption{A high quality blended level generated by the \textit{full} blend model trained off of World 9-3.}
  \label{fig:fullBlend}
\end{figure*}
Figure \ref{fig:world91Box} summarizes the results of the evaluation for World 9-1. While 9-1 is made up of a combination of ``overworld" and ``underwater" level sprites, it is much more overworld then underwater with a 6:1 ratio of sprites from each type. The models reflect this, with the Underwater level type model doing very poorly at explaining the level, while the SMB and Overworld level type models behave essentially the same. Despite this low quality blend, the blended model's distribution differs significantly from the SMB Model distribution according to the paired Wilcoxon-Mann-Whitney test ($p$=0.03327). In addition the blended model and full blended model distributions do not differ significantly ($p>0.05$), indicating that the system's choice for L nodes to blend is as good as creating all possible blended L nodes in this case. It is worth noting that the SMB model typically finds median scores for actual Super Mario Bros. levels between  0.1 and 0.2, with the lowest median score for any level being 0.05. None of these models reaches even the lowest point, but we contend this is due to the fact that the level does not represent a strong blend.

Figure \ref{fig:world93Box} summarizes the results of this evaluation for World 9-3. World 9-3 represents a much stronger blend than World 9-1 with an overworld to castle sprite ratio of 3:1. This is reflected in the relative distributions of the Castle and Overworld level type models. Once again the blended model distribution differs significantly from the SMB Model distribution ($p$=0.0008308). In this case the full blended model also differs significantly from the blended model ($p$=0.002961). However, despite the overall higher distribution, the full blended model's median value rose only a small amount compared to the blended model's median (0.077 vs 0.074). The full blended model is also made up of over two-hundred L nodes as opposed to our system's blended model of twenty-four L nodes. We therefore contend that our system picked out the most important L nodes to blend. In addition, both blended models' distributions fell into the range of an actual Super Mario Bros. level. We contend this is due to the level being a more even blend, indicating that our blending technique leads to blended models close in quality to those models trained directly on exemplar levels.

\subsection{Example Output}
We present a set of illustrative generated levels from our system. To create full levels our system determines the sequence of L nodes that best explains the sequence of uniform chunks of a target level. Each L node in this sequence is then prompted to generate a novel level chunk and the sequence of generated chunks constitutes a level. Figure \ref{fig:uwLvl} and Figure \ref{fig:high} represent high-quality levels (according to our system) using a blending target of World 9-1 and 9-3 respectively. In comparison we present Figure \ref{fig:mid} representing a lower quality blended level, and Figure \ref{fig:fullBlend} representing a high-quality level generated by the \textit{full} blend model. The difference between the low and high scoring levels should be clear from their structure, with Figure \ref{fig:mid} including individual, oddly placed blocks and a floating castle. We further identify a lack of difference between the blend and \textit{full} blend models, with Figure \ref{fig:high} and \ref{fig:fullBlend} appearing very similar.

The generated levels in Figure \ref{fig:uwLvl} and \ref{fig:high} demonstrate the quality of the blended models, but they are not perfect. For example, about three-fourths through Figure \ref{fig:high} there's a chunk where lava replaced ground. With additional knowledge this could have been avoided in the concept mapping phase. One element of future work will be attempting to learn properties of sprites from the gameplay video and integrating this knowledge into the blending process.

\section{Conclusions}
In this paper we've presented techniques to learn probabilistic models from gameplay video and to blend these models to produce novel level types. We ran a human subjects study to evaluate our model's ability to capture level design style as a measure of structural likelihood. We found strong evidence for this in the form of a strong correlation between participant's ranking of style and our system's rankings. We demonstrated via two case studies that our system is able to explain human expert blended levels, and is able to blend models that evaluate these levels significantly better than the unblended models. Taken together, these represent a system that is able to learn about design, evaluate design like a human, and is able to extend this knowledge to explain new domains via concept blending. Beyond improving the current blending process between level models, we also look toward blending models between multiple games in our future work. 

\section{Acknowledgements}
This material is based upon work supported by the National Science Foundation under Grant No. IIS-1525967. Any opinions, findings, and conclusions or recommendations expressed in this material are those of the author(s) and do not necessarily reflect the views of the National Science Foundation.

\bibliographystyle{iccc}
\bibliography{iccc}
\end{document}